\documentclass[journal]{IEEEtran}
\usepackage{cite}
\usepackage{amsmath,amssymb,amsfonts}
\usepackage{algorithmic}
\usepackage{graphicx}
\usepackage{textcomp}
\usepackage{xcolor}

\ifCLASSINFOpdf
\else
\fi
\begin{document}
%
\title{Learning to enhance multi-legged robot on rugged landscapes}
%
%
%

\author{Juntao He$^{1}$, Baxi Chong$^{2}$, Zhaochen Xu$^{2}$, Sehoon Ha$^{3}$, Daniel I. Goldman$^{2}$
\thanks{$^{1}$ J. He is with the Institute for Robotics and Intelligent Machines, Georgia Institute of Technology, Atlanta, USA. {\tt\small \{jhe391\}@gatech.edu}}
\thanks{$^{2}$ B. Chong, Z. Xu and D. Goldman are with School of Physics, Georgia Institute of Technology, USA.
            {\tt\small \{bchong9, 
            zxu699\}@gatech.edu, {\tt\small daniel.goldman@physics.gatech.edu}}} 
\thanks{$^{3}$ S. Ha is with the School of Computing, Georgia Institute of Technology, 
Atlanta, USA.
            {\tt\small \{sehoonha\}@gatech.edu}
    }}

\maketitle

\begin{abstract}
Navigating rugged landscapes poses significant challenges for legged locomotion. Multi-legged robots (those with 6 and greater) offer a promising solution for such terrains, largely due to their inherent high static stability, resulting from a low center of mass and wide base of support. Such systems require minimal effort to maintain balance. Recent studies have shown that a linear controller, which modulates the vertical body undulation of a multi-legged robot in response to shifts in terrain roughness, can ensure reliable mobility on challenging terrains. However, the potential of a learning-based control framework that adjusts multiple parameters to address terrain heterogeneity remains underexplored. We posit that the development of an experimentally validated physics-based simulator for this robot can rapidly advance capabilities by allowing wide parameter space exploration. Here we develop a MuJoCo-based simulator tailored to this robotic platform and use the simulation to develop a reinforcement learning-based control framework that dynamically adjusts horizontal and vertical body undulation, and limb stepping in real-time. Our approach improves robot performance in simulation, laboratory experiments, and outdoor tests. Notably, our real-world experiments reveal that the learning-based controller achieves a 30\% to 50\% increase in speed compared to a linear controller, which only modulates vertical body waves. We hypothesize that the superior performance of the learning-based controller arises from its ability to adjust multiple parameters simultaneously, including limb stepping, horizontal body wave, and vertical body wave.
\end{abstract}


%
\IEEEpeerreviewmaketitle

\section{Introduction}
\begin{figure}
    \centering
    \includegraphics[width=9cm]{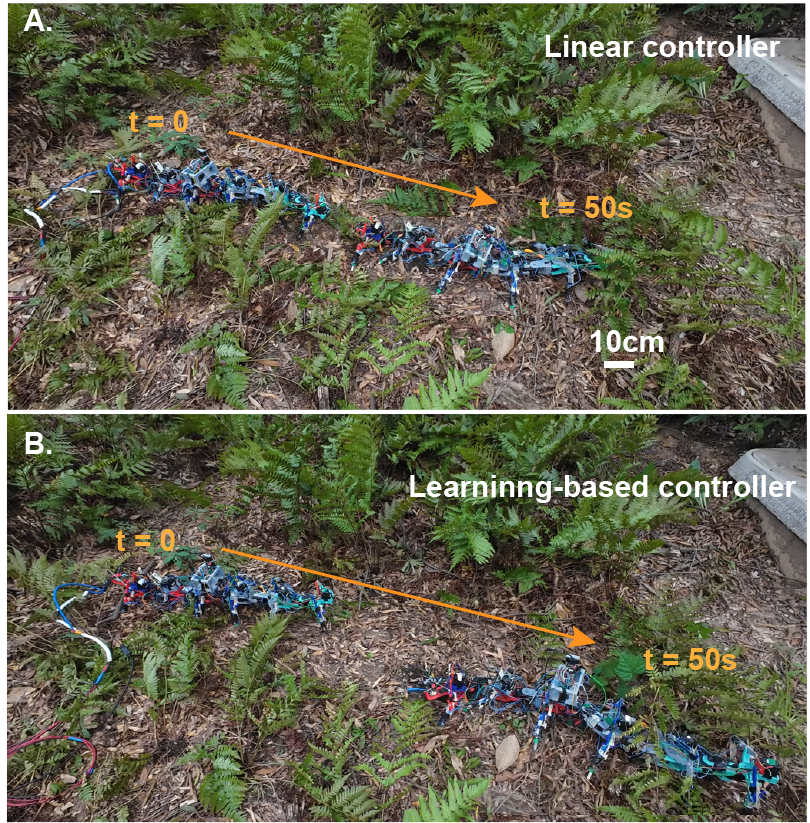}
    \caption{Outdoor experiments demonstrate a significant improvement in a multi-legged robot's speed by implementing a learning-based controller. On terrain composed of a mixture of bush, fern, and pine straw, the learning-based controller achieves a 50\% increase in speed compared to the linear controller.}
    \label{fig:1}
\end{figure}
%
%
%
%
Locomotion on highly rugged landscapes poses a significant challenge for legged robots. Few-legged robotic systems, such as bipedal and quadrupedal robots, have shown considerable mobility on rough terrain by leveraging advanced control frameworks, including Model Predictive Control (MPC) \cite{grandia2019feedback,bjelonic2021whole,farshidian2017real,grandia2021multi} and reinforcement learning-based control \cite{hwangbo2019learning,miki2022learning,rudin2022learning,yang2020data}. Multi-legged robots, typically equipped with more than six legs \cite{chong2023multilegged,chong2023self,ozkan2020systematic,he2024tactile}, present another promising solution for locomotion on challenging terrain.

Unlike few-legged systems, multi-legged robots are known for their high static stability \cite{ijspeert2023integration}, which is characterized by a low center of mass and a wide base of support. As a result, multi-legged robots do not need to exert significant effort to maintain balance on uneven terrain, unlike their fewer-legged counterparts. In recent studies, Chong et al. \cite{chong2023multilegged} demonstrated that a multi-legged robot can successfully navigate from point A to point B on rugged landscapes without the need for sensing, provided it has a sufficient number of legs (more than 10). However, the speed of robots using this approach tends to be relatively slow.

To address this, He et al. \cite{he2024tactile,he2024control} proposed a control framework (Fig. \ref{fig:4}.A) that enhances the speed of multi-legged robots on rugged landscapes by modulating vertical body undulation waves based on the approximation of terrain roughness. Specifically, the degree of terrain roughness is quantified by the discrepancy between the actual and ideal ground-foot contact states. Using this control framework, their robot achieves around 50\% improvement on robot's forward speed compared to previous open-loop control.  However, evidence \cite{chong2023self,chong2022general} suggests that further improvements could be made by also modulating the leg stepping wave and horizontal body undulation wave. This implies that extending the current single-input single-output controller to a multi-input multi-output framework could enhance the robot's speed even further.

In this work, we extended the linear controller proposed by He et al. \cite{he2024control,he2024tactile} by incorporating reinforcement learning. First, we created a MuJoCo-based multi-legged robot simulator, which we validated through real-world experiments. Subsequently, we developed a learning-based controller (Fig.\ref{fig:4}.B) that uses the amplitudes of the leg stepping wave, and horizontal and vertical body undulation waves, along with the ground-foot contact state as inputs during the current motion cycle to predict the optimal coordination of the three amplitudes for the next cycle. Finally, we demonstrated the effectiveness of our learning-based controller by testing our robot on both artificial laboratory rough terrain and natural outdoor rugged landscapes. Our results show that the learning-based controller improves the robot's speed by around 50\% compared to the linear controller.

\section{Related work}

\subsection{Wave patterns in multi-legged robot}
\label{wave pattern}
Previous research \cite{chong2022general,chong2023multilegged,chong2023self,he2024tactile} demonstrated that successful traversal of a multi-legged robot on rough terrain is achieved by coordinating leg movements with both horizontal and vertical body undulations (Fig.\ref{fig:4}). Specifically, forward motion is WAS generated by prescribing the limb stepping and body undulation as three sinusoidal traveling waves. 

Legs propel the robot at core by retracting during the stance phase to establish ground contact and protracting during the swing phase to disengage. Specifically, during the stance phase, the leg moves from the anterior to the posterior end, and reverses during the swing phase. To define the anterior/posterior excursion angles ($\theta_{leg}$) for a given contact phase ($\tau_c$), we use a piecewise sinusoidal function: 
\begin{align}
        \theta_{leg,l}(\tau_c,1)  &=\begin{cases}
      \Theta_{leg}\cos{(\frac{\tau_c}{2D})}, & \text{if}\ \text{mod}(\tau_c,2\pi)  < 2\pi D\\
      -\Theta_{leg}\cos{(\frac{\tau_c-2\pi D}{2(1-D)})}, & \text{otherwise},
    \end{cases} \nonumber \\
    \theta_{leg,l}(\tau_c, i) &= \theta_l(\tau_c - 2\pi\frac{\xi}{n}(i-1), 1) \nonumber \\
    \theta_{leg,r}(\tau_c, i) &= \theta_l(\tau_c + \pi, i) 
    \label{eq:legmove}
\end{align}

\noindent Here, $\Theta_{leg}$ is the amplitude of the shoulder angle, and $\theta_{leg,l}(\tau_c,i)$ and $\theta_{leg,r}(\tau_c,i)$ are the shoulder angles of the $i$-th left and right leg at the contact phase $\tau_c$. The shoulder angle reaches its maximum ($\theta_{leg}=\Theta_{leg}$) during the transition from swing to stance and its minimum ($\theta_{leg}=-\Theta_{leg}$) during the reverse transition. We assume $D=0.5$ unless specified otherwise.

Lateral body undulation is introduced by propagating a wave from head to tail: \begin{align} \theta_{body}(\tau_b,i)=\Theta_{body} \cos(\tau_b - 2\pi\frac{\xi^b}{n}(i-1)), \label{eq
} \end{align}

\noindent where $\theta_{body}(\tau_b,i)$ is the angle of the $i$-th body joint at phase $\tau_b$, and $\xi^b$ denotes the number of spatial waves on the body. We assume an equal number of spatial waves in both body undulation and leg movement ($\xi^b = \xi$), allowing lateral body undulation to be prescribed by its phase $\tau_b$.

The gaits of multi-legged locomotors, characterized by the phases of contact ($\phi_c$) and lateral body undulation ($\tau_b$), are achieved through the superposition of a body wave and a leg wave. Optimal body-leg coordination, ensuring effective leg retraction, is given by $\phi_c=\tau_b-(\xi/N+1/2)\pi$.

Vertical body undulation is introduced by propagating a wave along the backbone: \begin{align} \theta_v(\tau_b,i)=A_v \cos(2\tau_b - 4\pi\frac{\xi^b}{n}(i-1)), \label{eq
} \end{align} \noindent where $\theta_v(\tau_b,i)$ is the vertical angle of the $i$-th body joint at phase $\tau_b$, and $A_v$ specifies its amplitude.

\subsection{Feedback control for multi-legged robot}
\label{Section linear controller}
Designing a feedback control framework for a multi-legged robot on rough terrain presents significant challenges due to the robot’s over 25 degrees of freedom (DOF) and its complex interactions with the environment. Recent studies \cite{he2024tactile,he2024control} introduced a linear controller that modulates vertical body undulation in response to variations in terrain roughness. These studies demonstrated that the robot's speed is closely linked to the ground-foot contact state, measured by prismatic contact sensors located at the distal-most segments of the limbs. The findings show that minimizing the discrepancy between the actual and ideal ground-foot contact significantly improves locomotion speed.

As in He et al., we quantify the relationship between ground-foot contact and locomotion speed using a parameter called the contact ratio ($\beta$), which measures the likelihood of actual contact aligning with the ideal contact state. A contact ratio of $\beta = 1$ indicates perfect contact, while values closer to 0 reflect significant disturbances. We found that the robot's speed is approximately linearly correlated with the contact ratio, with higher $\beta$ values leading to faster movement.

Furthermore, modulating the amplitude of vertical body undulation can reduce disturbances to the contact state caused by environmental variability. Consequently, we implemented a linear controller (Fig. \ref{fig:4}.A) that adjusts the vertical wave amplitude ($A_v$) based on the discrepancy between the actual and ideal ground-foot contact states, significantly improving the robot's speed. This linear controller is described by the equation: 
\begin{equation} 
A_v = K_p (\beta_s - \beta_0), 
\label{eq: linear controller} \end{equation} 
where $K_p$ is the proportional gain of the controller, $\beta_s$ is the actual contact ratio measured by the sensors, and $\beta_0$ is the expected contact ratio.

Real-world experiments validated this approach, showing a 30\% to 60\% improvement in speed compared to an open-loop controller. 
\begin{figure}
    \centering
    \includegraphics[width=9cm]{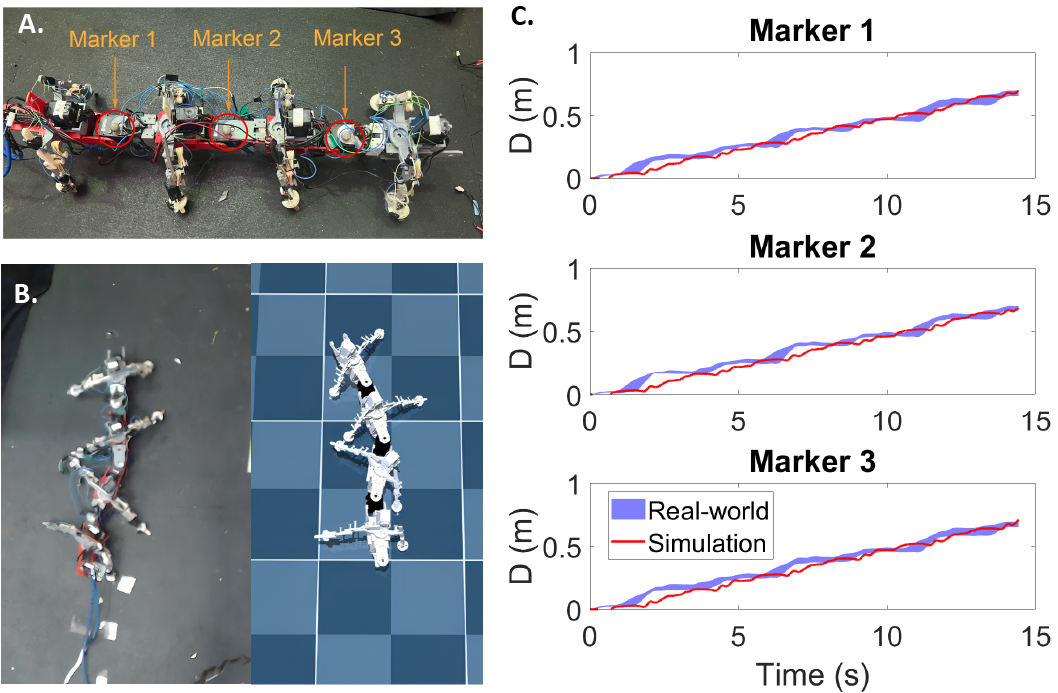}
    \caption{\textbf{Simulation validation on Flat Ground.} A. Marker assembly locations on the robot. B. Snapshots from both the real-world experiment and the simulation.  C. This figure illustrates the displacement over time for three markers, comparing the results from both the simulation and the real-world experiments.}
    \label{fig:2}
\end{figure}
\subsection{Reinforcement learning for legged locomotion}
Reinforcement learning (RL) has proven effective in training agile and dynamic motor skills for legged robots \cite{xie2023learning,kumar2023cascaded,yokoyama2023asc,hwangbo2019learning,miki2022learning}. With trained RL policies, quadrupedal robots have demonstrated impressive mobility on challenging terrains such as snow, mud, wet moss, and rocks \cite{hwangbo2019learning,miki2022learning,rudin2022learning,smith2022legged}. Additionally, RL has shown effectiveness in training policies that coordinate robotic arm manipulation and locomotion skills for humanoid robots and quadrupedal robots equipped with robotic arms \cite{yokoyama2023asc,cheng2024expressive,li2023ace}.

In this paper, we aim to address this gap by training a multi-legged robot using reinforcement learning. Specifically, we will develop a learning-based control policy that predicts the optimal coordination of three wave patterns (Section \ref{wave pattern}) to maximize the robot's speed on rough terrain.

\section{Method}
We first detail the development of MuJoCo-based simulators for our multi-legged robot, followed by real-world experiments to verify the reliability of these simulators. We then present the formulation of our learning-based controller.
\begin{figure}[ht]
    \centering
    \includegraphics[width=9cm]{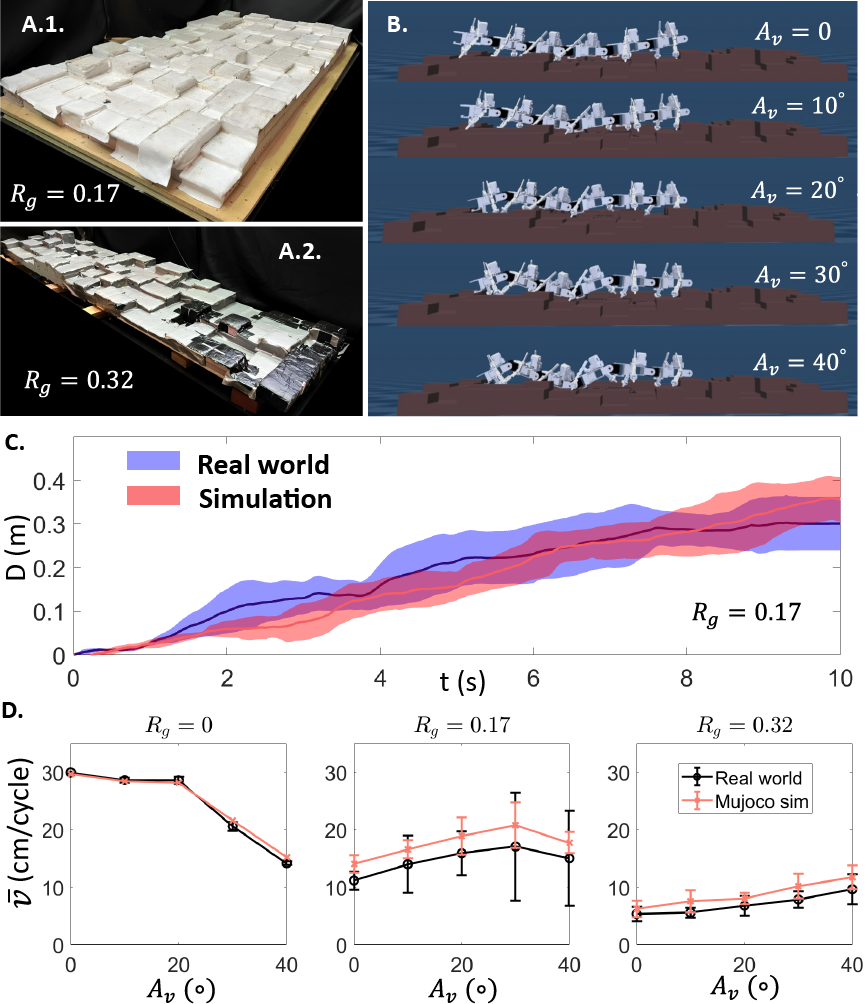}
    \caption{\textbf{Simulation validation on rough terrain.} A. Terrain with different roughness. B. Screenshot depicting the robot moving forward with varying vertical amplitudes ($A_v$) on the $R_g=0.17$ terrain. C. Displacement versus time for robot moving on $R_g=0.17$ terrain with $A_v=20^{\circ}$. D.Velocity versus vertical amplitude plot comparing simulation and real-world experiments.}
    \label{fig:3}
\end{figure}
\subsection{Multi-legged robot simulator}

To generate data for the reinforcement learning algorithm and validate the policy through simulation, we developed MuJoCo-based simulators for the multi-legged robot. MuJoCo, a free and open-source physics engine, is designed to support research and development in fields such as robotics, biomechanics, graphics, animation, and other areas where fast and accurate simulation is essential. To demonstrate the reliability of our simulators, we compared real-world experiments with simulations to assess the sim-to-world gap.

For the initial validation on flat terrain, we conducted three motion cycles of an 8-legged robot on a level surface. The physical robot's motion was tracked using an Optitrak motion capture system, and the collected real-world data was then compared to the corresponding data generated by the MuJoCo simulation. As illustrated in Figure.\ref{fig:2}, the simulation and real-world experiment show a strong alignment in displacement history. Specifically, the Root Mean Square Error (RMSE) across three markers, averaged over the trials, is less than 3.5 cm.

We then expanded our validation efforts by testing the robot on laboratory rough terrains, specifically the two terrains depicted in Fig. \ref{fig:3}.A. These terrains consist of 10 cm by 10 cm blocks, with heights that follow a normal distribution and are controlled by the parameter $R_g$. The parameter $R_g$, referred as rugosity, quantifies the roughness of a bumpy terrain, where a higher $R_g$ indicates a more rugged surface. The correlation between $R_g$ and the standard deviation (std) of the height distribution is given by std = $12.5R_g$ cm. Additional details on the construction of the rough terrains can be found in \cite{chong2023multilegged,he2024tactile}. Next, we tested the robot on terrains with varying levels of roughness ($R_g$) and different amplitudes of vertical body undulation ($A_v$) (Fig. \ref{fig:3}.B). In each experimental trial on flat ground and $R_g=0.32$ terrain, we executed three motion cycles and recorded the average speed per cycle. On $R_g=0.17$ terrain, we ran the robot for two cycles of motion. As shown in Fig. \ref{fig:3}.B, the simulation results closely align with the outcomes obtained from real-world experiments. Specifically, on flat ground, terrains with $R_g=0.17$ and $R_g=0.32$, the discrepancies between the simulation and real-world data are less than 3\%, 12\% and 8\% respectively.
\begin{figure*}
    \centering
    \includegraphics[width=18cm]{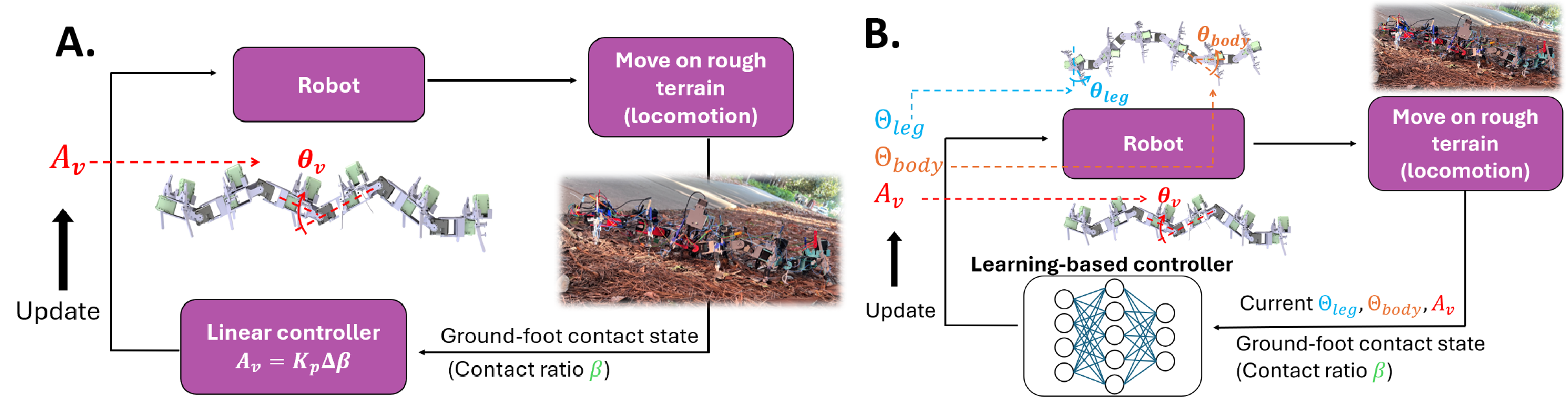}
    \caption{\textbf{Control frameworks for the linear and learning-based controllers. A. Linear controller:} This controller modulates the vertical body undulation wave based on real-time ground-foot contact data from the sensors (contact ratio $\beta$). Here $A_v$ represents the amplitude of the vertical body wave, and $\theta_v$ denotes the joint angle of the vertical body joint. The parameter $K_p$ refers to the proportional gain of the linear controller, and $\Delta \beta$ indicates the discrepancy between the actual and expected contact ratios.\textbf{B. Learning-based controller:} This controller adjusts limb stepping and both horizontal and vertical body undulation waves, based on the amplitudes of these three waves alongside real-time ground-foot contact ratio ($\beta$). In this case, $\Theta_{leg}$ and  $\Theta_{body}$ correspond to the amplitudes of the leg and body waves, respectively, while $\theta_{leg}$ and $\theta_{body}$ represent the leg joint angle and the horizontal body angle, respectively. 
}
    \label{fig:4}
\end{figure*}
\subsection{Learning-based control policy}

\subsubsection{Background: Markov Decision Process}
Reinforcement learning can be modeled as a Markov Decision Process (MDP), defined by the 5-tuple $\mathcal{M} = \langle S, A, P, R, \gamma \rangle$, where $S$ represents the state space, $A$ the action space, and $R$ the reward function. The transition function $P$ predicts the next state given the current action. The discount factor, $\gamma \in (0,1)$, adjusts the importance of future rewards. The objective is to find the optimal policy $\pi: S \rightarrow A$ that maximizes the cumulative return, $V^{\pi}(s) = \mathbb{E}_{\tau \sim \pi} \left[\sum_{t=0}^{T} R(s_t, a_t)\right]$, where $\tau$ denotes the distribution of trajectories generated by the policy $\pi$.

\subsubsection{Learning formulation}

The current linear controller (Fig.\ref{fig:4}.A) and Eq.\ref{eq: linear controller} for the multi-legged robot modulates the vertical body wave amplitudes ($A_v$) while keeping the leg wave and horizontal body wave fixed, based on the real-time ground-foot contact ratio ($\beta$) (Section \ref{Section linear controller}). However, the robot's performance can benefit from dynamically adjusting both the leg wave and horizontal body wave. Our previous study\cite{chong2022general,chong2023self} found that increasing the horizontal body wave amplitude ($\Theta_{body}$) and leg wave amplitude ($\Theta_{leg}$) could enhance the robot's speed on flat ground. However, on rough terrain, high values of either $\Theta_{body}$ or $\Theta_{leg}$ can cause significant yaw instability. 

Therefore, a control policy that can properly coordinate the three wave amplitudes ($A_v$, $\Theta_{body}$, $\Theta_{leg}$) is highly needed for more stable locomotion. Using reinforcement learning (RL), we trained a learning-based controller (Fig.\ref{fig:4}.B) that outputs the optimal coordination of $A_v$, $\Theta_{body}$, and $\Theta_{leg}$ based on the current wave amplitudes and the ground-foot contact state. 

The state space defines inputs to a policy, which includes the amplitudes ($A_v$, $\Theta_{body}$, $\Theta_{leg}$) of the leg stepping wave, horizontal and vertical body undulation waves, and the ground-foot contact ratio ($\beta$). 

The learning objective is to learn an effective controller, which maximizes the robot's forward speed while minimizing lateral displacement caused by yaw motion. To achieve this, we define our reward function as follows:
\begin{equation}
    R = v_{f} - 0.6  \mid v_{l} \mid,
\end{equation}
where $v_f$ represents the forward speed while $v_l$ represents the lateral speed.
The discount factor $\gamma$ is set to 0.99 for all the experiments.

During training, we set each step as one full motion cycle of the robot, which lasts 3 seconds in the simulation. This allows the robot to complete the entire sequence of leg stepping and body undulation waves. The advantage of this setup is that the policy can converge quickly, resulting in smooth robot movement. However, a potential drawback is that the policy might converge to local optima within the given constraints.

\subsubsection{Domain Randomization}

\begin{figure}
    \centering
    \includegraphics[width=9cm]{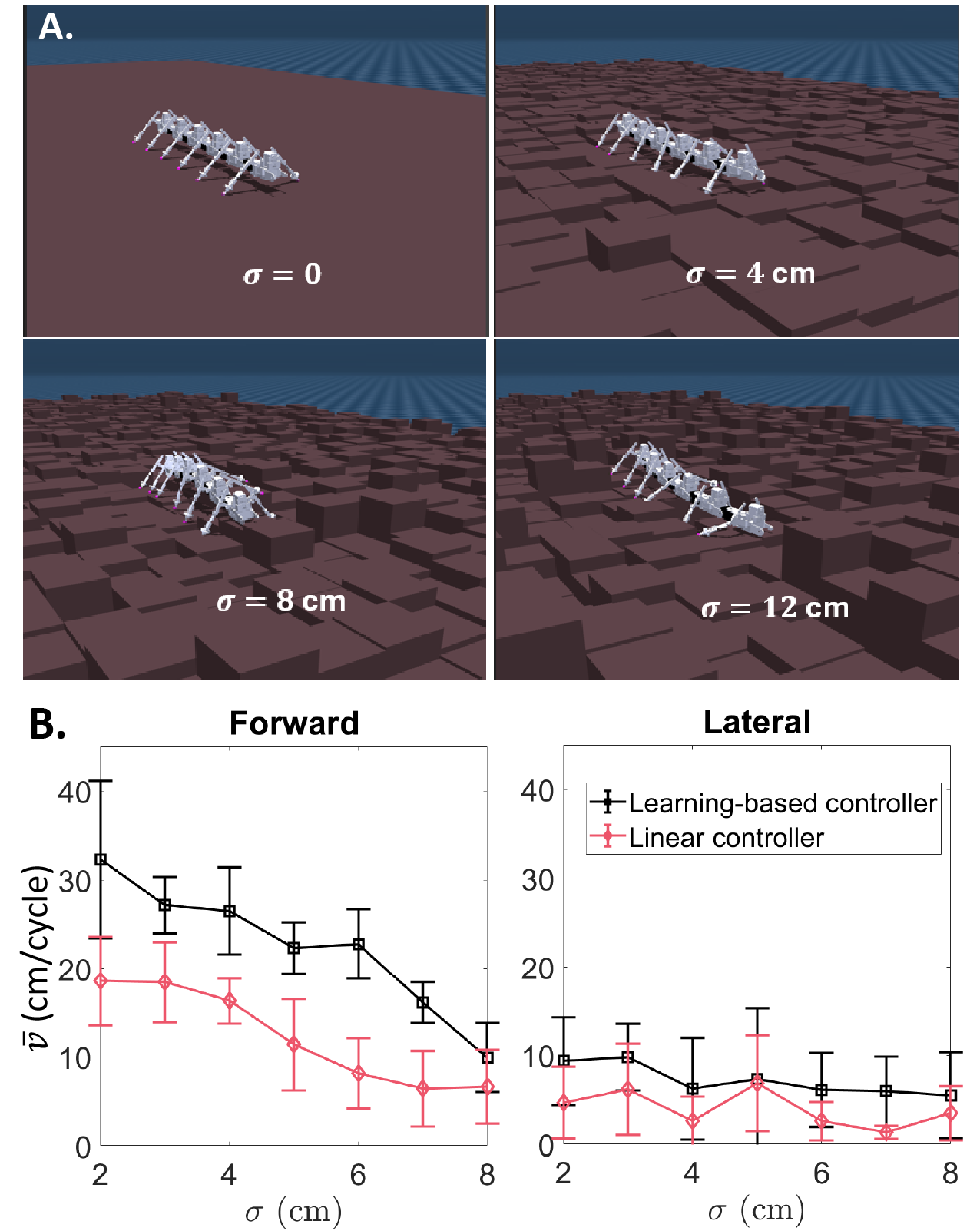}
    \caption{\textbf{Rough terrain variation in simulation and simulation results.} \textbf{A.} The roughness of the terrain in simulation is varied by adjusting the parameter $\sigma$, which modifies the standard deviation of block heights, effectively randomizing the terrain conditions. \textbf{B.} The simulation results present a comparison between the performance of the learning-based controller and the linear controller. The average speed per cycle, denoted as $\bar{v}$, is used as the performance metric.
    }
    \label{fig:5}
\end{figure}

To reduce the sim-to-real gap, we adjust the terrain roughness during training (Fig.\ref{fig:5}.A) to ensure our policy is robust across a broader range of rough terrains. In the simulation, we create a 10 m $\times$ 3 m step field composed of 10 cm $\times$ 10 cm blocks with varying heights. The heights of these blocks follow a normal distribution, with a mean of 20 cm and a variance of $\sigma$ centimeters. The value of $\sigma$ is randomly selected in the simulation and reset every 16 steps.

\subsubsection{Implementation Details}

In our work, we use PPO (Proximal Policy Optimization) \cite{schulman2017proximal} as the  RL algorithm to train our policy. PPO is a reinforcement learning algorithm designed to improve policy optimization by balancing performance and stability. It is an on-policy method that updates the policy using a surrogate objective function, which helps prevent large and potentially destabilizing updates. For more details on its implementation, please refer to the original paper \cite{schulman2017proximal}.


The standard objective function of PPO is expressed as:
\begin{equation}
    L^{\text{CLIP}}(\theta) = \hat{\mathbb{E}}_t \left[ \min \left( r_t(\theta) \hat{A}_t, \text{clip}(r_t(\theta), 1 - \epsilon, 1 + \epsilon) \hat{A}_t \right) \right],
\end{equation}
where $\theta$ represents the policy parameters, $\hat{\mathbb{E}}_t$ denotes the empirical expectation over timesteps, $r_t(\theta)$ is the ratio of probabilities under the new and old policies, $\hat{A}_t$ is the estimated advantage at time $t$, and $\epsilon$ is a hyperparameter. In our training, we set $\epsilon = 0.2$.

\section{Results}
We next demonstrate the effectiveness of our learning-based controller through a combination of simulation and real-world experiments. The real-world tests were carried out on rough terrains in both controlled laboratory environments and outdoor field conditions. During these experiments, the robot was operated using both a linear controller and the learning-based controller, allowing us to directly compare their performance and quantify the improvements achieved by the learning-based approach.
\subsection{Simulation results}
We first evaluated the learning-based policy using the MuJoCo simulator, running the robot across rough terrains with $\sigma$ values ranging from 2 cm to 8 cm. As shown in the simulation results in Fig. \ref{fig:5}.B, the learning-based controller improves forward speed by 30\% to 60\% without causing significant lateral displacement. We also observed that the performance of both controllers declines sharply as $\sigma$ approaches 8 cm. This may be because the height difference between adjacent blocks becomes so large that it acts as a wall, obstructing the robot’s movement.
\subsection{Lab-based test}
\begin{figure}
    \centering
    \includegraphics[width=9cm]{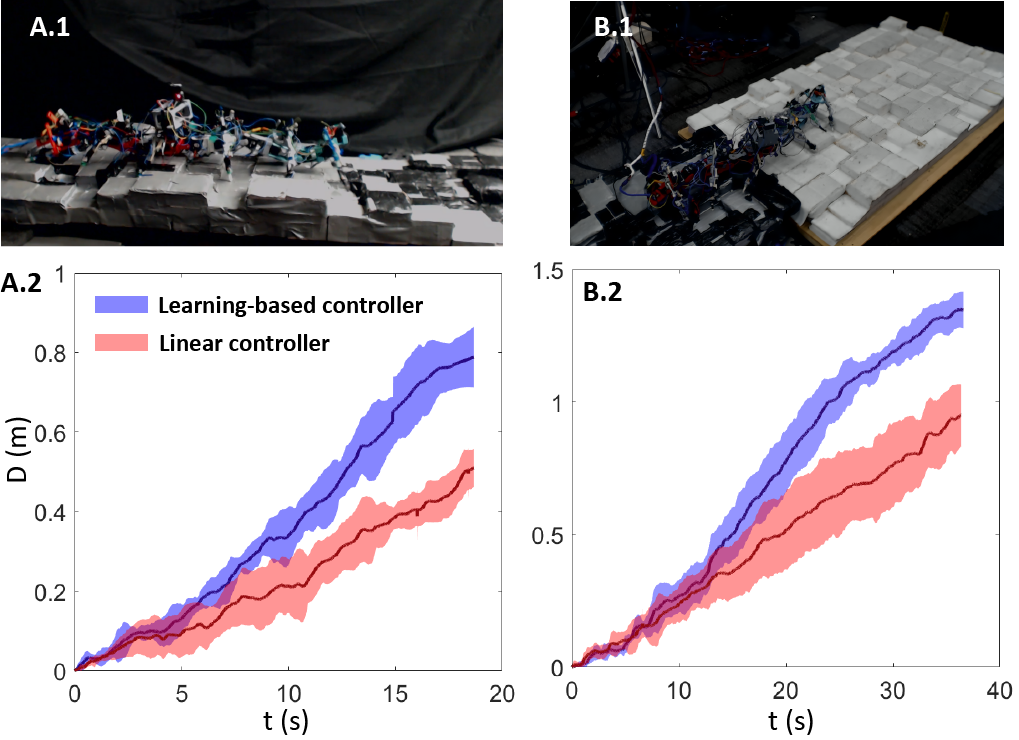}
    \caption{\textbf{Controller comparision in laboratory experiments} \textbf{A.1 and B.1:} Snapshots showing the robot moving on rough terrains. In A.1, $R_g = 0.32$. In B.1, a combination of $R_g = 0.17$ and $R_g = 0.32$ is used. \textbf{A.2 and B.2:} Displacement versus time, comparing the performance of the learning-based controller with the linear controller.}
    \label{fig:6}
\end{figure}
The first lab-based terrain we tested is a rough terrain with $R_g = 0.32$ (Fig. \ref{fig:6}.A.1). We ran the robot on this terrain for 4 motion cycles using both the linear controller and the learning-based controller, with a motion capture system (Opti-Track) tracking the robot's displacement over time. As shown in Fig.\ref{fig:6}.A.2, the learning-based controller improved the robot's forward speed by approximately 60\%.

Next, we tested the robot on a terrain combining $R_g = 0.17$ and $R_g = 0.32$ (Fig. \ref{fig:6}.B.1). On this terrain, the robot was run for 8 motion cycles with both controllers, and its displacement was again tracked by the motion capture system. The results in Fig. \ref{fig:6}.B.2 indicate that the learning-based controller increased the robot's speed by approximately 50\%.

\subsection{Outdoor test}
\begin{figure}
    \centering
    \includegraphics[width=9cm]{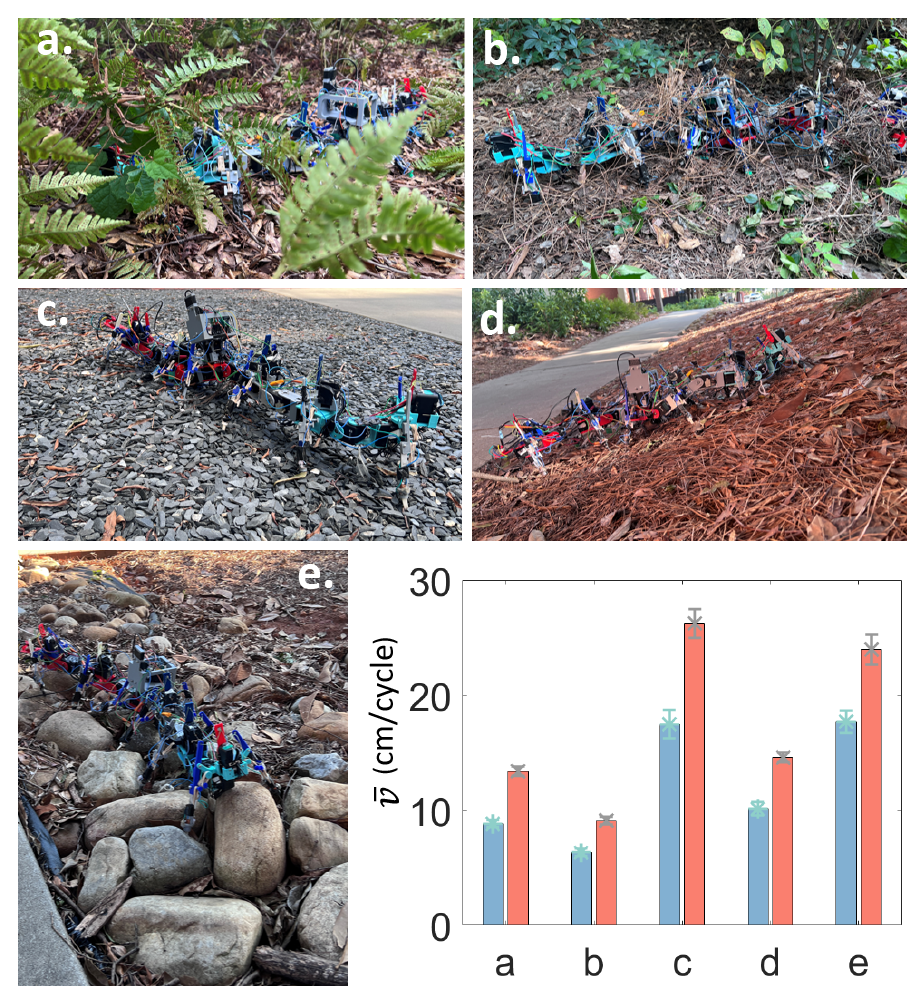}
    \caption{\textbf{Testing controllers in diverse outdoor terrains.} \textbf{Pictures of the robot in wild:} Collection of rugged landscapes. a. Terrain with a mixture of fern, pine straw, mud, and bushes. b. Terrain with a mixture of bushes, mud, and heavy pine straw. c. Terrain with a mixture of mud and giant boulders. d. Terrain with rocks. e. Terrain with a mixture of even ground, mud, and heavy pine straw. The slope is $20^\circ$. \textbf{Bottom right corner:} Forward speed data from experiments conducted across all five outdoor terrains. $\bar{v}$ represents the average speed per cycle. The x-axis corresponds to the terrain index shown in the left picture. The learning-based controller, shown in red, consistently outperforms the linear controller (in blue) by achieving higher average speeds across all terrains.}
    \label{fig:7}
\end{figure}

To evaluate the performance of the learning-based controller under more challenging conditions, we conducted a series of field tests with our robot in outdoor environments. As shown in Fig.\ref{fig:7}, the tests were carried out on five rugged terrains. Terrain (a) consists of a mixture of fern, pine straw, mud, and bushes. Terrain (b) features a combination of bushes, mud, and heavy pine straw. Terrain (c) is primarily rocky, while Terrain (d) is a mix of even ground, mud, and heavy pine straw, with a slope of $20^{\circ}$. Terrain (e) is characterized by mud interspersed with large boulders. 

To compute the average speed (displacement per cycle), we measured the robot's displacement using a tape measure and divided this displacement by the total number of motion cycles. Our measurements, as illustrated in Fig. \ref{fig:7}, indicate that the learning-based controller increased the robot's speed by approximately 30\% to 60\%. 
\section{Conclusion}
In this work, we extended the linear controller introduced by He et al. \cite{he2024control,he2024tactile} by integrating reinforcement learning to develop a more sophisticated, learning-based control strategy for the  multi-legged robot. Our approach leverages the amplitudes of the leg stepping wave, horizontal and vertical body undulation waves, and the ground-foot contact state as inputs within each motion cycle, enabling the prediction of optimal amplitude coordination for subsequent cycles. To support this development, we created a MuJoCo-based multi-legged robot simulator, which was validated through comprehensive real-world experiments. The effectiveness of our learning-based controller was demonstrated through tests conducted on both artificial indoor rough terrain and natural outdoor rugged landscapes. The results indicate that our approach improves the robot's speed by approximately 50\% compared to the original linear controller.

The current work constrains leg stepping and body undulation by coupling the joints into three sinusoidal waves. In the future, we plan to relax this constraint during training to explore whether it can further enhance the performance of the learning-based controller. At present, the robot relies solely on contact sensors on each foot to understand its interaction with the environment. We also intend to integrate additional sensors, such as head-mounted collision detection sensors, into the learning framework to assess whether the robot can benefit from this additional information. Furthermore, we plan to extend this learning framework to train the robot for traversing sloped and rough terrains.


%





\ifCLASSOPTIONcaptionsoff
  \newpage
\fi

\bibliographystyle{IEEEtran}
\bibliography{bibtex/bib/IEEEabrv,bibtex/bib/IEEEexample}

\end{document}